\documentclass[conference]{IEEEtran}
\IEEEoverridecommandlockouts
\usepackage[utf8]{inputenc} % Cho phép nhập ký tự Unicode trực tiếp
\usepackage[T5]{fontenc}    % Bảng mã font T5 chuẩn cho tiếng Việt
\usepackage{booktabs}
\usepackage{cite}
\usepackage{amsmath,amssymb,amsfonts}
\usepackage{tabularx}
\usepackage{graphicx}
\usepackage{textcomp}
\usepackage{xcolor}
\usepackage{float}
\usepackage{balance}
\usepackage[font=footnotesize,labelfont=bf]{caption}
\usepackage{hyperref} % For hyperlinks
\hypersetup{colorlinks=true, linkcolor=black, citecolor=black, urlcolor=black} % IEEE thường dùng màu đen cho link khi in

\DeclareCaptionLabelFormat{IEEE}{#1~#2}
\def\BibTeX{{\rm B\kern-.05em{\sc i\kern-.025em b}\kern-.08em
    T\kern-.1667em\lower.7ex\hbox{E}\kern-.125emX}}

\begin{document}

\title{From Benchmarking to Reasoning: 
A~Dual-Aspect, Large-Scale Evaluation of LLMs on Vietnamese Legal Text\thanks{Accepted at the FISU Joint Conference on Artificial Intelligence (FJCAI 2026), Vietnam.}}

\author{\IEEEauthorblockN{Van-Truong Le}
\IEEEauthorblockA{\textit{University of Science, VNUHCM}\\
\textit{Vietnam National University} \\
\textit{Hanoi Open University}\\
Ho Chi Minh, Vietnam \\
23120181@student.hcmus.edu.vn}
}

\maketitle

\begin{abstract}
The complexity of Vietnam's legal texts presents a significant barrier to public access to justice. While Large Language Models offer a promising solution for legal text simplification, evaluating their true capabilities requires a multifaceted approach that goes beyond surface-level metrics. This paper introduces a comprehensive dual-aspect evaluation framework to address this need. First, we establish a performance benchmark for four state-of-the-art large language models (GPT-4o, Claude 3 Opus, Gemini 1.5 Pro, and Grok-1) across three key dimensions: Accuracy, Readability, and Consistency. Second, to understand the "why" behind these performance scores, we conduct a large-scale error analysis on a curated dataset of 60 complex Vietnamese legal articles, using a novel, expert-validated error typology. Our results reveal a crucial trade-off: models like Grok-1 excel in Readability and Consistency but compromise on fine-grained legal Accuracy, while models like Claude 3 Opus achieve high Accuracy scores that mask a significant number of subtle but critical reasoning errors. The error analysis pinpoints \textit{Incorrect Example} and \textit{Misinterpretation} as the most prevalent failures, confirming that the primary challenge for current LLMs is not summarization but controlled, accurate legal reasoning. By integrating a quantitative benchmark with a qualitative deep dive, our work provides a holistic and actionable assessment of LLMs for legal applications.
\end{abstract}


\section{Introduction}
The advent of Large Language Models (LLMs) offers a transformative promise: to democratize access to justice by translating complex, codified legislation into language the public can understand. In civil law systems like Vietnam's, this promise is particularly profound. The nation's laws, while comprehensive, are often shrouded in ``legalese''—a dense legal terminology that creates a significant barrier for citizens seeking to understand their fundamental rights and obligations \cite{nicholson2010access}. LLMs present a potential tool to remove this barrier.

However, this promise is shadowed by a significant danger: the risk of generating fluent, plausible, yet inaccurate legal simplifications \cite{dahl2024large}. To harness the potential of LLMs responsibly, we must first be able to measure their capabilities with nuance and depth. Initial evaluations, including our own preliminary work, have focused on setting performance benchmarks using surface-level metrics such as legal accuracy, user-perceived readability, and output consistency \cite{kelsall2025rapid}. Although these metrics provide a valuable ``what''—quantifying which models perform better on the surface—they fundamentally fail to explain the ``why.'' A model might achieve a high accuracy score by correctly summarizing the general rule for inheritance, yet completely miss a critical exception for a specific circumstance, a subtle but catastrophic reasoning error that superficial scores would mask. This limitation was a key piece of feedback on our initial research, which was criticized for its small sample size and lack of novel insight into the models' failure modes.

This highlights a critical gap in the literature: the absence of a holistic evaluation framework that marries large-scale quantitative benchmarking with deep qualitative diagnostics of legal reasoning \cite{kelsall2025rapid}. To fill this gap, this paper introduces a comprehensive, dual-aspect evaluation framework designed to be both broad and deep. We combine two distinct but complementary research thrusts:
\begin{enumerate}
    \item \textbf{A Large-Scale Performance Benchmark:} We expand upon our previous methodology, applying the metrics of Accuracy, Readability, and Consistency to a large and diverse dataset of 60 complex Vietnamese legal articles to ensure the generalizability and statistical significance of our findings.
    \item \textbf{An In-Depth Legal Reasoning Error Analysis:} We introduce and apply a novel, expert-validated error typology to the same outputs. This allows us to move beyond scores, dissecting each model's performance to identify and categorize the specific root causes of their reasoning failures.
\end{enumerate}

By integrating these quantitative and qualitative approaches, we provide more than just a leaderboard of models. Our work offers a detailed diagnostic of the current capabilities and, more importantly, the systemic weaknesses of state-of-the-art LLMs in the Vietnamese legal domain. The findings reveal crucial trade-offs between performance aspects—such as readability versus factual fidelity—and pinpoint the core challenges that must be overcome to build truly reliable and trustworthy legal AI.

\section{Related Work}
The application of LLMs in the legal domain has rapidly evolved, yet significant gaps remain, particularly for non-English languages and civil law systems. Our work is situated within several key research streams.

\subsection{Domain Adaptation and Pre-trained Legal Models}
A foundational line of research demonstrates that domain-specific pre-training significantly improves performance on legal tasks. Models like LEGAL-BERT \cite{chalkidis2020legal} show that continued pre-training on large legal corpora enables models to better understand legal vocabulary and context compared to general-purpose models. This highlights the importance of domain-specific data, a challenge for languages like Vietnamese where large, digitized legal corpora are less abundant.

\subsection{Legal Text Simplification}
Simplifying legal texts is a recognized task aimed at improving access to justice. Research in this area has explored various techniques, from unsupervised methods combining lexical replacement and sentence splitting \cite{cemri2022unsupervised} to supervised approaches. However, a recurring challenge, as noted by Garimella et al. \cite{garimella2022text}, is the scarcity of parallel complex-simple legal corpora, making supervised training difficult. This scarcity elevates the importance of evaluating the zero-shot simplification capabilities of modern LLMs, which is a focus of our study.

\subsection{Evaluation and Reliability of Legal LLMs}
As LLMs become more powerful, evaluating their reliability in high-stakes domains is paramount. Surveys like Lai et al.~\cite{lai2024large} provide a broad overview of LLM applications in law, from drafting assistance to legal Q\&A, while also outlining risks related to reasoning, transparency, and factual accuracy. A critical aspect of evaluation is understanding and mitigating ``hallucinations.'' Dahl et al.~\cite{dahl2024large}, for instance, developed a typology for legal hallucinations focused primarily on factual inaccuracies and citation errors in case law summarization and legal Q\&A tasks.

While such work is foundational for identifying fact-based errors, our research addresses a different, yet equally critical, challenge: \textbf{reasoning failures within the context of text simplification}. Our nine-category error typology is specifically designed to capture nuances unique to this task. Unlike typologies focused on hallucinations (i.e., fabricating non-existent facts), our framework also categorizes subtle but critical errors like \textit{Oversimplification}, \textit{Misinterpretation} of legal terms, and failure to apply principles correctly in \textit{Incorrect Examples}. This focus on simplification-specific reasoning errors, rather than just factual correctness, represents a key contribution that complements existing evaluation frameworks. It underscores the inadequacy of surface-level accuracy metrics and motivates our deep dive into error analysis.

\subsection{Vietnamese Legal Language Processing}
In the context of Vietnamese, efforts to develop specialized legal AI resources are accelerating. On the research front, projects like ViGPT-Law \cite{pham2025top} and legal Q\&A datasets such as VLQA and VNLAWQC \cite{nguyen2025vlqa} are building foundational models and benchmarks, primarily for information retrieval and question answering. In parallel, the Vietnamese government has deployed practical tools, such as the Ministry of Justice's ``Virtual Legal Assistant,'' aimed at improving public access to legal information.

However, a critical gap exists between creating these applications and fundamentally understanding their reliability. While existing efforts often emphasize factual correctness, they may overlook subtle but critical reasoning failures in generative tasks like simplification. Our work directly addresses this gap. While applied tools provide valuable services, our research offers a fundamental, diagnostic evaluation of the underlying LLM capabilities, identifying systemic failure modes through a novel error typology. This large-scale, systematic evaluation of generative simplification, viewed through the lens of legal reasoning, is a contribution that complements existing work. It not only advances technical benchmarks but also provides an actionable framework for auditing and improving the safety and reliability of public-facing legal AI systems in Vietnam.

\section{A Dual-Aspect Evaluation Framework}
To provide a holistic assessment, we designed a framework that evaluates LLM performance from two complementary perspectives: high-level performance metrics and low-level error analysis. The entire framework was applied to a large dataset to ensure statistical significance and generalizability, while also enabling a deeper understanding of how different models succeed or fail across various aspects of legal reasoning.

\subsection{Data and Model Selection}
\textbf{Models:} We selected four state-of-the-art LLMs representing the current forefront of commercial AI: \textbf{GPT-4o} (OpenAI), \textbf{Claude 3 Opus} (Anthropic), \textbf{Gemini 1.5 Pro} (Google), and \textbf{Grok-1} (xAI). These models were chosen based on three criteria: (1) Market Dominance: They represent the flagship models from the leading AI laboratories; (2) Architectural Diversity: They embody different training philosophies and context window capabilities (e.g., Gemini’s long context vs. Grok’s MoE architecture); and (3) Accessibility: They are currently the most accessible high-performance tools for general users, making their evaluation highly relevant for public legal access.

\textbf{Dataset:} We curated a dataset of 60 legal articles via purposive sampling. This large-scale approach prioritizes both depth and breadth. The articles were selected from three core legislative texts: the Penal Code 2015 (20 articles), the Civil Code 2015 (20 articles), and the Land Law 2024 (20 articles). The selection criteria focused on articles known for their complexity, including those with: (a) multiple exceptions and conditions, (b) abstract legal terminology (e.g., ``trái đạo đức xã hội''), and (c) requirements for sequential logical reasoning (e.g., ``thừa kế thế vị''). This ensures a challenging and representative testbed for the LLMs. By combining civil, criminal, and land law, the dataset reflects a wide range of legal reasoning challenges encountered in both daily life and specialized legal contexts. Moreover, this carefully balanced corpus provides sufficient variation to stress-test the adaptability of different model families under comparable conditions.

\textbf{Task:} For each article, models were given a consistent, zero-shot prompt asking them to act as a legal assistant and explain the law in simple terms with a practical example for a layperson. The specific prompt used was: ``Act as a legal assistant. Explain the following article [Article Content] in simple terms for a layperson and provide a practical example''. Each model generated two responses per article (at temperature 0.2), creating a corpus of 480 outputs ($4$ models $\times$ $60$ articles $\times$ $2$ runs) for evaluation. This experimental design enables us to evaluate not only the accuracy and accessibility of the generated simplifications but also the stability of model behavior across repeated generations. Notably, this dual-component prompt—requiring both explanation (a summarization task) and example generation (an application task)—allowed us to observe that models consistently struggled more with the latter, highlighting a core challenge in legal reasoning over mere text simplification.

\subsection{Phase 1: Overall Performance Benchmarking}
This phase quantifies the overall quality of the generated texts across three user-centric dimensions, providing a high-level comparison of the models.

\subsubsection{Metric 1: Legal Accuracy}
This metric assessed the legal precision of the outputs. It was rated by a team of five trained law students on a 1-5 Likert scale. To ensure reliability, each output was rated by at least two students, and any significant discrepancies were resolved through discussion. The metric is a weighted average of four sub-criteria:
\begin{itemize}
    \item \textbf{Content Preservation (40\%):} Fidelity to the core legal rule.
    \item \textbf{Completeness (30\%):} Inclusion of all essential elements and conditions.
    \item \textbf{Clarity (20\%):} How easy the legal definition is to understand.
    \item \textbf{Example Relevance (10\%):} Appropriateness of the provided example.
\end{itemize}
The weighting scheme was established based on the principle of `Safety First' in legal advice. Content Preservation is assigned the highest weight (40\%) because a simplification that omits core legal meaning poses the greatest risk of misinformation. Completeness (30\%) follows, ensuring no critical conditions are missed. Clarity (20\%) and Example Relevance (10\%) are weighted lower, as a legally accurate but dry explanation is preferable to a fluent but incorrect one.

\subsubsection{Metric 2: Readability}
This metric assessed the ease of understanding for a non-expert audience. It was rated by a diverse group of 253 non-expert participants recruited via university mailing lists and social media. Each participant rated a random subset of 10-15 outputs on a 1-5 Likert scale across three weighted sub-criteria:
\begin{itemize}
    \item \textbf{Ease of Language (40\%):} Simplicity of vocabulary and sentence structure.
    \item \textbf{Structural Coherence (30\%):} Logical flow, formatting, and organization.
    \item \textbf{Utility of Examples (30\%):} How helpful the example was in clarifying the concept.
\end{itemize}
For readability, Ease of Language is weighted highest (40\%) as it directly addresses the barrier of `legalese' for laypeople, followed equally by Structural Coherence (30\%) and Utility of Examples (30\%), which support comprehension and retention.

\subsubsection{Metric 3: Consistency}
This metric measured the stability of a model's output across two independent runs. This composite metric combined automated scores and human judgment:
\begin{itemize}
    \item \textbf{Semantic Consistency (30\%):} Cosine similarity between sentence embeddings (using \texttt{paraphrase-multilingual-MiniLM-L12-v2}).
    \item \textbf{Linguistic Consistency (30\%):} Sentence-level BLEU score to measure phrasing overlap.
    \item \textbf{Length Consistency (10\%):} Stability in output length (word count).
    \item \textbf{Manual Consistency Score (30\%):} Human evaluator rating of meaning preservation across runs.
\end{itemize}
Length Consistency (10\%): While output length is not a direct measure of reasoning quality, significant variance in length for identical prompts (e.g., producing a brief summary in one run and a detailed essay in another) indicates system instability. This metric serves as a proxy for the model's determinism and reliability in a production environment.

\subsection{Phase 2: In-Depth Legal Reasoning Error Analysis}
This phase moves beyond scores to identify the root causes of failure.

\subsubsection{Error Typology} We developed a nine-category error typology, validated and refined in collaboration with legal experts from Hanoi Open University. This framework, detailed in Table \ref{tab:error_typology}, allows for a consistent and granular assessment of model outputs. The typology is designed to capture a wide range of potential failures, from simple omissions to complex reasoning flaws.

\subsubsection{Annotation Process} The same team of five law students who rated for accuracy also performed the error annotation. To ensure high-quality annotations, a rigorous two-stage process was employed. First, two raters independently annotated each of the 480 outputs. The initial inter-rater agreement was measured using Cohen's Kappa, yielding a score of \textbf{$\kappa = 0.72$}. This value indicates substantial agreement between the annotators and validates the clarity of our error categories. Second, all discrepancies were flagged and resolved in a group session moderated by the lead researcher to produce the annotated dataset of annotated errors.

% Figure 2
\begin{figure*}[t!]
    \centering
    % Chỉnh width=0.8\textwidth để ảnh chiếm 80% chiều rộng trang
    \includegraphics[width=0.8\textwidth]{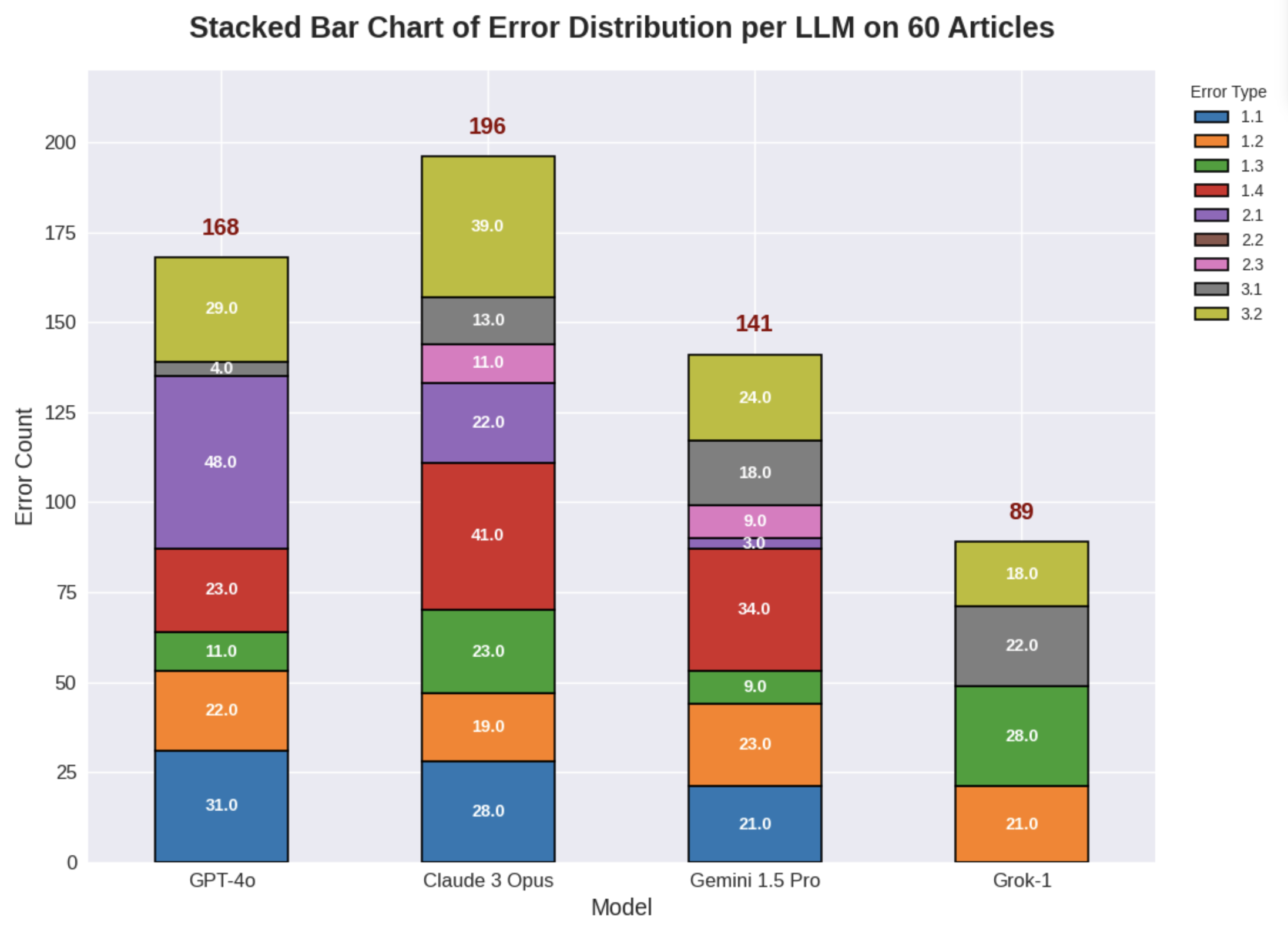} 
    \caption{Stacked Bar Chart of Error Distribution per LLM on 60 Articles. The chart clearly shows Grok-1's lower total error count and its unique profile, which lacks errors in categories 1.1, 1.4, and 2.1.} 
    \label{fig:stacked_bar_scaled}
\end{figure*}

% Bảng 1 rộng, dùng table* để trải 2 cột
\begin{table*}[htbp]
\centering
\caption{The Nine Categories of the Legal Reasoning Error Typology}
\label{tab:error_typology}
\begin{tabularx}{\linewidth}{>{\centering\arraybackslash}p{0.08\linewidth} 
                               >{\raggedright\arraybackslash}p{0.25\linewidth} 
                               >{\raggedright\arraybackslash}X}
\toprule
\textbf{ID} & \textbf{Error Name} & \textbf{Definition} \\
\midrule
1.1 & Omission of Core Elements & Fails to mention a core condition, subject, right, or obligation. \\
1.2 & Omission of Exceptions & Fails to mention exceptions or limitations. \\
1.3 & Hallucination & Fabricates information or legal consequences. \\
1.4 & Misinterpretation & Incorrectly explains the meaning of a legal term. \\
2.1 & Oversimplification & Simplifies to the point of losing critical legal nuance. \\
2.2 & Cross-reference Failure & Fails to incorporate info from a referenced article. \\
2.3 & Internal Contradiction & Provides contradictory statements in the same response. \\
3.1 & Irrelevant Example & Provides an example that does not match the provision. \\
3.2 & Incorrect Example & Provides a relevant example but draws a legally incorrect conclusion. \\
\bottomrule
\end{tabularx}
\end{table*}

\section{Experiments and Results}
Our dual-aspect evaluation yielded a rich dataset, allowing us to both rank the models on performance and diagnose their underlying weaknesses.

\subsection{Phase 1: Overall Performance Results}
The large-scale benchmark revealed clear trade-offs between the models' capabilities. The aggregated scores, presented in Table \ref{tab:performance_scores}, provide a high-level view of their strengths.

\begin{table}[htbp]
\centering
\caption{Overall Performance Scores (Scaled 1--5)}
\label{tab:performance_scores}
\begin{tabularx}{\linewidth}{l *{4}{>{\centering\arraybackslash}X}}
\toprule
\textbf{Metric} & \textbf{GPT-4o} & \textbf{Claude 3} & \textbf{Gemini 1.5} & \textbf{Grok-1} \\
\midrule
Accuracy & 3.78 & \textbf{4.71} & 4.65 & 4.39 \\
Readability & 4.55 & 4.78 & 4.41 & \textbf{4.93} \\
Consistency & 3.49 & 3.81 & 3.94 & \textbf{4.28} \\
\midrule
\textbf{Total} & 4.02 & 4.55 & 4.42 & \textbf{4.57} \\
\bottomrule
\end{tabularx}
\end{table}

Interestingly, while Claude 3 Opus had a slightly lower total score, Grok-1 emerged as the top performer by a narrow margin when all factors were considered. \textbf{Grok-1} demonstrated a compelling profile, dominating in both Readability and Consistency. \textbf{Claude 3 Opus} was the clear leader in Legal Accuracy but was held back by slightly lower scores in other areas. \textbf{Gemini 1.5 Pro} showed strong accuracy but was significantly hampered by poor readability, often producing large, unstructured blocks of text. \textbf{GPT-4o} consistently trailed the other models.

% Figure 1
\begin{figure}[htbp]
\centering
\includegraphics[width=\columnwidth]{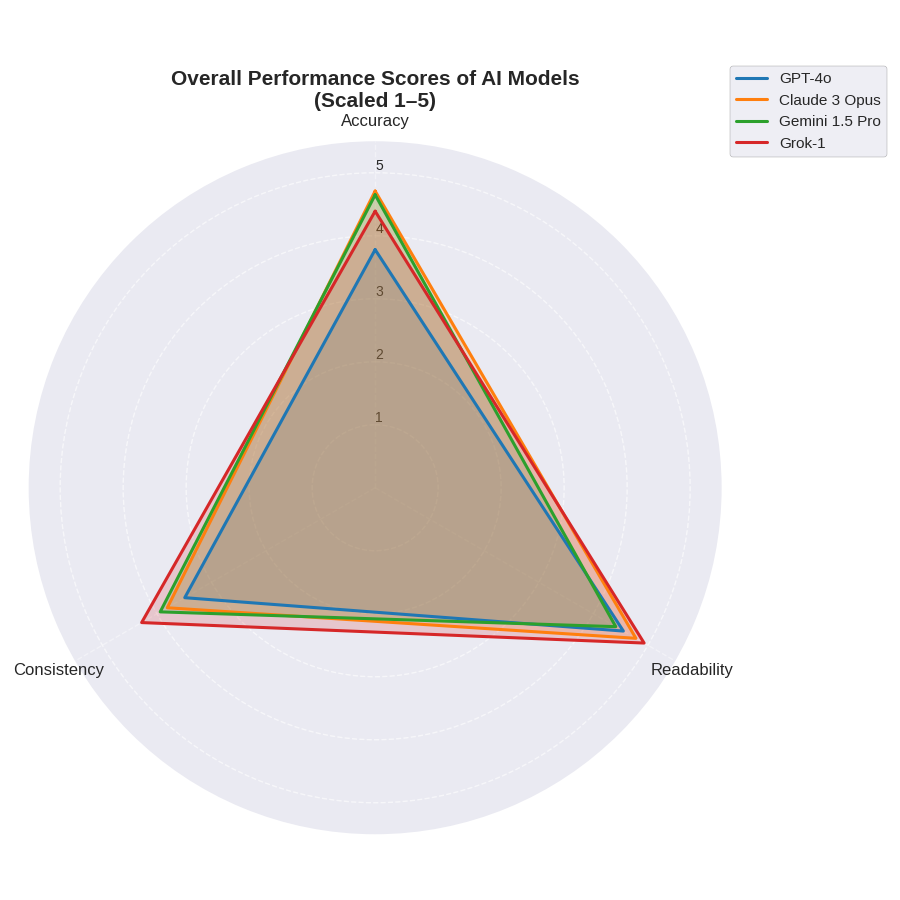} 
\caption{Radar Chart of Overall Performance Scores. The chart visualizes the distinct strengths of each model: Claude 3 Opus's strength in Accuracy, Grok-1's dominance in Readability and Consistency, and Gemini's imbalance.}
\label{fig:radar_performance}
\end{figure}

\subsection{Phase 2: Legal Reasoning Error Analysis Results}
The error analysis, summarized in Table \ref{tab:error_matrix_scaled}, provides a deeper understanding of the scores from Phase 1.

% Bảng 3 cũng rất rộng (10 cột), dùng table*
\begin{table*}[htbp]
\centering
\caption{Error Frequency Matrix by Model and Error Type (based on 480 responses)}
\label{tab:error_matrix_scaled}
\scriptsize % Giảm font cho vừa
\begin{tabularx}{\linewidth}{l *{10}{>{\centering\arraybackslash}X}}
\toprule
\textbf{Model} & \textbf{1.1} & \textbf{1.2} & \textbf{1.3} & \textbf{1.4} & \textbf{2.1} & \textbf{2.2} & \textbf{2.3} & \textbf{3.1} & \textbf{3.2} & \textbf{Total} \\
\midrule
GPT-4o          & 31 & 22 & 11 & 23 & 48 & 0 & 0 & 4  & 29 & \textbf{168} \\
Claude 3 Opus  & 28 & 19 & 23 & 41 & 22 & 0 & 11 & 13 & 39 & \textbf{196} \\
Gemini 1.5 Pro & 21 & 23 & 9  & 34 & 3  & 0 & 9  & 18 & 24 & \textbf{141} \\
Grok-1         & 0  & 21 & 28 & 0  & 0  & 0 & 0  & 22 & 18 & \textbf{89}  \\
\bottomrule
\end{tabularx}
\end{table*}

The large-scale data reinforce the distinct ``personalities'' of each model:

\paragraph{Grok-1: The Cautious Complier} With the lowest error count by a significant margin (89), Grok-1 proved to be the safest model. It committed zero errors of \textit{Omission of Core Elements}, \textit{Misinterpretation}, or \textit{Oversimplification}. Its failures were almost exclusively concentrated in generative tasks, such as fabricating details in examples (Error 1.3) or providing irrelevant analogies (Error 3.1).

\paragraph{Claude 3 Opus: The Ambitious but Risky Lawyer} Claude 3 Opus consistently attempted the most sophisticated analyses and, as a result, made the most nuanced and severe errors (196 total). Its high frequency of \textit{Misinterpretation} errors (41) was particularly alarming. It repeatedly failed on articles involving subtle legal concepts, such as the distinction between different types of invalid civil transactions or the precise requirements for ``urgent situations.'' These tendencies indicate that Claude 3 Opus aims high but often overreaches in ways that compromise reliability.

\paragraph{GPT-4o: The Eager but Oversimplifying Teacher} GPT-4o's defining flaw, observed consistently, was \textit{Oversimplification} (48 errors). In numerous instances, it reduced complex, multi-condition rules into simplistic, often dangerously misleading, maxims. This highlights GPT-4o’s difficulty in balancing accessibility with fidelity to the original legal text.

\paragraph{Gemini 1.5 Pro: The Inconsistent Performer} Gemini's performance was marked by high variance. It produced many perfect, error-free responses but also exhibited significant logical failures, including instances of \textit{Internal Contradiction} and a high number of \textit{Irrelevant Examples}. Such inconsistency makes Gemini less predictable and therefore harder to trust in sensitive legal contexts.

\subsection{Synthesis: Connecting Performance Metrics to Error Types}
By combining both phases, we can move beyond a simple leaderboard to draw a richer set of conclusions about the models' underlying behaviors.

\begin{itemize}
    \item \textbf{The ``Accuracy'' Illusion:} Claude 3 Opus's top score in Accuracy is misleading when viewed in isolation. Its high score is achieved by correctly handling the simpler components of legal articles, which masks a high propensity to fail on the most complex parts. Its leading count of critical \textit{Misinterpretation} errors reveals a fragility in abstract reasoning that aggregate scores alone cannot capture.
    \item \textbf{The Readability vs. Fidelity Trade-off:} Grok-1's top scores in Readability and Consistency are a direct result of its ``cautious,'' source-adherent nature. This strategy ensures high stability but hampers its ability to generate novel, accurate applications of the law. Its primary failures—\textit{Irrelevant} and \textit{Incorrect Examples}—demonstrate a clear trade-off: it excels at faithful paraphrasing at the cost of being a less capable reasoner.
    \item \textbf{Oversimplification as a Double-Edged Sword:} GPT-4o's performance demonstrates how prioritizing accessibility can backfire. Its dominant error, \textit{Oversimplification}, is the very mechanism it uses to achieve readable outputs. By stripping away essential legal nuances for the sake of simplicity, it creates distortions that undermine legal accuracy and render its simplifications potentially harmful.
\end{itemize}

\section{Discussion}
Our large-scale findings provide robust evidence that current LLMs, while fluent, still struggle systematically with the core demands of legal reasoning. The consistent prevalence of \textit{Incorrect Example} and \textit{Misinterpretation} errors across a broad dataset confirms that models are significantly more capable at rephrasing existing text than at applying legal principles to novel scenarios. This highlights a fundamental gap between linguistic competence and abstract reasoning capabilities in the legal domain.

\subsection{Impact of Model Architecture and Alignment Strategies}
The distinct error profiles observed in Phase 2 suggest that performance differences are driven not just by model size, but by underlying architectural choices and alignment philosophies.

\textit{Grok-1 and the ``Alignment Tax'' Hypothesis:}
Grok-1's unexpected dominance in \textit{Readability} and \textit{Consistency}, coupled with its lack of critical interpretation errors, can be attributed to its unique design. As a massive Mixture-of-Experts (MoE) model with 314 billion parameters, Grok-1 possesses a high capacity for nuance. More critically, unlike models from OpenAI or Anthropic which undergo aggressive Reinforcement Learning from Human Feedback (RLHF) focused on safety and refusal, Grok-1 appears to operate with looser alignment constraints. We hypothesize that Grok-1 suffers less from the ``alignment tax''—a phenomenon where excessive safety tuning degrades performance on specific downstream tasks. This allows it to adhere faithfully to the simplification instruction without over-filtering or sanitizing the output, resulting in a ``cautious complier'' behavior that prioritizes fidelity to the source text over risky inferential leaps.

\textit{The Paradox of Safety in GPT-4o:}
Conversely, GPT-4o's lower ranking reveals a tension between safety and legal nuance. Its primary failure mode, \textit{Oversimplification} (Error 2.1), likely stems from robust safety guardrails and instruction tuning that favor conciseness and neutrality. When processing complex penal or civil codes, GPT-4o exhibits a conservative behavior, stripping away necessary but intricate conditions to provide a safe, high-level summary. While this reduces the risk of generating specific hallucinations, it renders the simplification legally shallow and less useful for analyzing edge cases, effectively trading accuracy for safety.

\textit{Claude 3 Opus and Inference-Heavy Reasoning:}
Claude 3 Opus's performance reflects the ``helpfulness'' objective of its Constitutional AI (CAI) training. The model frequently attempts to make abstract connections to provide insightful explanations. While this ambition yields high accuracy on standard interpretations, it leads to a high rate of \textit{Misinterpretation} when the law requires strict literalism. This suggests that optimization for ``helpful'' conversation may inadvertently encourage the model to overreach, generating plausible but legally incorrect inferences.

\subsection{Contextualizing Findings within Broader Literature}
Our results corroborate and extend recent findings in Legal NLP. Similar to Dahl et al. \cite{dahl2024large}, who identified pervasive hallucinations in case law generation, we found that even when models have access to the correct statute, they struggle to apply it accurately (Error 3.2: \textit{Incorrect Example}). However, our study reveals a distinct failure mode specific to simplification: the \textbf{fluency-accuracy trade-off}. While Dahl et al. focused on factual fabrication, our findings suggest that models can generate highly fluent, logically sounding legal reasoning that contains subtle interpretation errors. This confirms that high linguistic competence often masks deficits in logical reasoning, creating an ``illusion of competence'' that is particularly risky for lay users.

\subsection{Practical Implications: Towards Risk-Aware Human-in-the-Loop}
The scale of this study strengthens the conclusion that deploying LLMs for public-facing legal aid without robust human oversight is premature. However, our error typology offers a blueprint for more efficient oversight. Rather than treating all outputs as equally suspect, a system could apply our findings as a risk-assessment framework. For example, outputs from ``inference-heavy'' models like Claude 3 Opus could be flagged for interpretation review, while example generation from all models should be routed to legal professionals, given the high failure rate in Error 3.2. This shifts human oversight from a passive safeguard to an active, risk-aware process.

\subsection{Limitations and Future Work}
While this study establishes a rigorous framework, we acknowledge specific methodological limitations that affect the interpretation and generalizability of our results.

First, regarding \textbf{data scale}, our dataset consists of 60 legal articles. Although these were purposively sampled to ensure complexity, this sample size is relatively small compared to the vastness of the Vietnamese legal corpus. However, this scale was necessary to ensure the depth and quality of our rigorous manual annotation process, which would be unfeasible with a larger dataset. Consequently, the findings may not fully capture the performance variability across other specialized legal domains, such as intellectual property or administrative law.

Second, regarding \textbf{evaluator expertise}, the reliance on senior law students rather than practicing attorneys is a limitation. While we mitigated this through rigorous training and achieved substantial inter-rater agreement ($\kappa=0.72$), we acknowledge that students may lack the practical experience required to detect the most subtle nuances in legal application errors compared to professional experts.

Third, the experimental design was restricted to a \textbf{zero-shot setting}. We deliberately chose this to benchmark the models' inherent, unassisted capabilities. However, this approach likely establishes a performance ``lower bound.'' We recognize that advanced techniques such as Few-Shot Learning or Chain-of-Thought (CoT) prompting could significantly mitigate specific reasoning failures, particularly \textit{Oversimplification}, which remains an open question for future research. Furthermore, this study represents a static snapshot of rapidly evolving models; newer iterations released after our evaluation period may exhibit different behaviors.

Finally, the study relies exclusively on commercial closed-source models. Future benchmarks should include open-weight models (e.g., Llama 3, Qwen) to enhance reproducibility and allow for a deeper investigation into how training data transparency correlates with legal reasoning performance.

\section{Conclusion}
This study provides the first large-scale, dual-aspect evaluation of leading LLMs on complex Vietnamese legal texts. By integrating a quantitative benchmark with a qualitative error analysis across 60 articles, we demonstrated that each model exhibits a unique and stable ``error profile.'' We found that no model is flawlessly reliable and that the most significant, consistent challenges lie in applying legal rules to new examples and correctly interpreting nuanced legal concepts. Our research contributes a reusable error typology and provides robust empirical evidence that, while promising, LLM technology requires significant advancements in reasoning and a framework of expert human oversight before it can be safely deployed to simplify law for the public. Ultimately, our dual-aspect framework serves as a critical blueprint for evaluating high-stakes AI applications not only in Vietnam but also in other civil law jurisdictions and low-resource languages facing similar challenges.

\section*{Acknowledgment}
We would like to express our sincere gratitude to the team of lecturers and law students from Hanoi Open University for their invaluable participation and expertise in the evaluation process. Their insights were crucial to the validation of the error typology and the annotation of the large-scale dataset.

\balance
\bibliographystyle{plain} % Kiểu hiển thị danh mục
\bibliography{references} % Tên file .bib (không cần .bib)
\vspace{12pt}

\end{document}